%% file: AerialDojo.tex
\documentclass{article} 
\usepackage{AerialDojo,times}
\input{math_commands.tex}

\usepackage{hyperref}
\usepackage{url}

\usepackage{amsmath,amssymb,booktabs,graphicx,array,longtable}
\usepackage{tabularx}
\usepackage[table]{xcolor}
\usepackage{subcaption}
\usepackage{wrapfig}
\usepackage{colortbl}
\usepackage{float}
\definecolor{lastrowblue}{HTML}{ECF4FF}
\newcolumntype{Y}{>{\raggedleft\arraybackslash}X}
\definecolor{expblue}{HTML}{ECF4FF}
\definecolor{expgroup}{HTML}{F3F5F8}

\title{AerialDojo-200K: A Large-Scale Benchmark Suite for Open-World Aerial Object-Goal Search}

\author{Tongtong Feng$^1$, Xin Wang$^{1*}$, Haoran Hou$^1$, Ren Wang$^1$, Weiran Wang$^2$, Shaokai Zhu$^3$, \\ \textbf{Ziqi Jia$^1$, Hao Wang$^1$, Yu-Wei Zhan$^4$, Zongyuan Wu$^1$, Jinghao Cui$^1$, Wenwu Zhu$^{1}$}\thanks{Corresponding Authors.} \\
$^1$\textnormal{Department of Computer Science and Technology, BNRist, Tsinghua University}\\
$^2$\textnormal{School of Electrical \& Electronic Engineering, University College Dublin}\\
$^3$\textnormal{School of Electronics Engineering and Computer Science, Peking University}\\
$^4$\textnormal{Department of Computer Science and Technology, Zhejiang University of Technology}\\
}

\iclrfinalcopy 

\begin{document}
\maketitle
\begin{abstract}
Open-world aerial object-goal search is a foundational yet challenging task, requiring aerial agents to autonomously explore large-scale, unstructured three-dimensional environments and reach target objects specified by semantic descriptions or reference images, rather than following route-specific instructions. However, research in this task remains at a nascent stage and relies on small, environment-specific benchmarks with heterogeneous action spaces and data formats. These limitations hinder large-scale training and cross-benchmark evaluation, constraining the scalability and generalizability of aerial agents. To address this problem, we propose AerialDojo-200K, a large-scale benchmark suite for open-world aerial object-goal search, with $3\times$ (times) as many scenes and $18.7\times$ as many task instances as the largest existing benchmark for this task. Specifically, we construct 42 simulation scenes spanning four scene families and 21 scene types, including 18 urban, 12 natural, six infrastructure, and six disaster scenes. To ensure data quality, 12 annotators spent two months manually annotating 109 landmarks, 2099 target objects, and 2099 object anchors across these scenes. We further construct 205,732 task instances, comprising over 100K semantic-goal and over 100K image-goal instances across Base, Standard, and Long-Horizon settings. Each task instance includes a collision-free reference trajectory and corresponding multi-view video recordings. We also develop a unified evaluation framework with a scene partition comprising 21 in-distribution scenes and 21 out-of-distribution scenes. Finally, our evaluation of five open-source and four closed-source multimodal large language models reveals that there is still a long way to go toward achieving general-purpose aerial agents. All can be found in  https://fengtt42.github.io/AerialDojo/.
\end{abstract}

\section{Introduction}
Open-world Aerial Object-Goal Search (AerialOGS)~\citep{uavon} requires aerial agents to autonomously search for target objects through goal-driven, long-horizon exploration in large-scale, unstructured three-dimensional environments, and to actively stop when onboard observations match the target object specified by a semantic description or reference image. Unlike aerial Vision-and-Language Navigation (VLN), which follows detailed, step-by-step route instructions that constrain how the aerial agent should travel~\citep{zhao2026worldvln}, AerialOGS provides only a high-level, open-ended object-goal and must jointly resolve where to explore, how to navigate, and when the observed object satisfies the goal. AerialOGS has broad potential applications in search and rescue, infrastructure inspection, and autonomous delivery, where operators may specify the object without knowing its exact location or providing a complete flight route.

\begin{figure}[t]
\centering
\includegraphics[width=\linewidth]{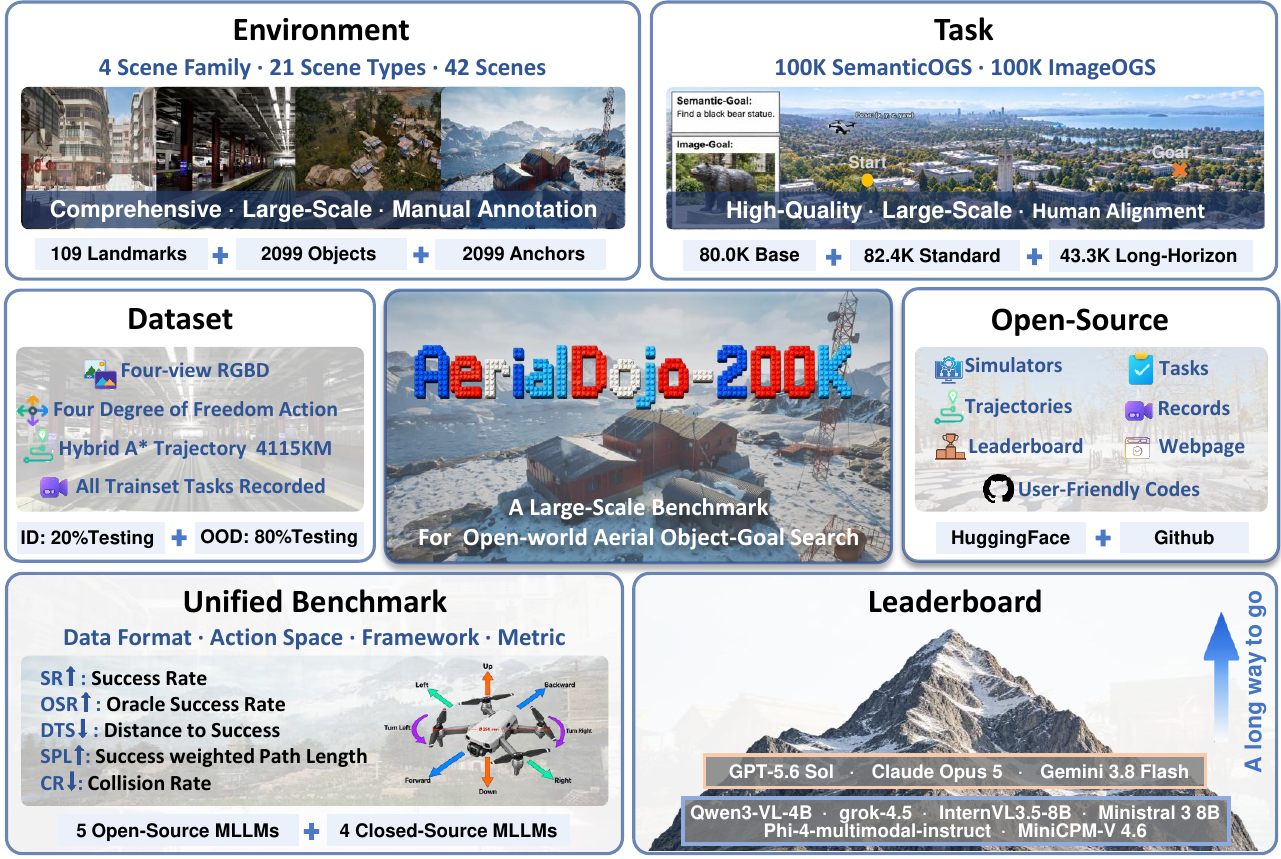}
\caption{The overview of AerialDojo-200K.}
\label{fig1}
\vspace{-0.4cm}
\end{figure}

Research on AerialOGS tasks remains in a nascent stage. On the one hand, existing aerial agents utilize spatio-semantic representations~\citep{zhang2026apex} to organize memories and plan over explored regions, frontiers, and object candidates. On the other hand, recent studies demonstrate that multimodal large models can leverage high-level semantic reasoning to interpret semantic goals~\citep{uavon} and guide search decisions~\citep{chen2026airhunt}. 

Nevertheless, the primary performance bottleneck remains the lack of large-scale benchmarks for training and evaluation. As shown in Table~\ref{tab1}, 1) UAV-ON \citep{uavon} and CityAVOS \citep{cityavos} are currently the only AerialOGS benchmarks, and the scale of those is approximately one-tenth as many tasks as OpenFly \citep{openfly}; 2) Heterogeneous action spaces and data formats across AerialOGS and VLN benchmarks make it difficult for existing VLN benchmarks to support AerialOGS tasks. Consequently, existing research relies on small, environment-specific benchmarks with heterogeneous action spaces and data formats, failing to support large-scale training and cross-benchmark evaluation, and limiting the scalability and generalizability of aerial agents.

To address this problem, we propose AerialDojo-200K, a large-scale benchmark suite for open-world aerial object-goal search, with $3\times$ (times) as many scenes and $18.7\times$ as many task instances as the largest existing benchmark for this task. Specifically, as shown in Figure~\ref{fig1}, we construct 42 simulation scenes spanning four scene families and 21 scene types, including 18 urban, 12 natural, six infrastructure, and six disaster scenes. To ensure data quality, 12 annotators spent two months manually annotating 109 landmarks, 2099 target objects, and 2099 object anchors across these scenes. We further construct 205,732 task instances, comprising over 100K semantic-goal and over 100K image-goal instances across Base, Standard, and Long-Horizon settings. Each task instance includes a collision-free reference trajectory and corresponding multi-view video recordings, with the trajectories totaling 4115km across the dataset. We also develop a unified evaluation framework with a scene partition comprising 21 in-distribution scenes and 21 out-of-distribution scenes. Finally, our evaluation of five open-source and four closed-source multimodal large language models reveals that there is still a long way to go toward achieving general-purpose aerial agents. Our main contributions are summarized as follows:

\begin{itemize}
    \item We propose AerialDojo-200K, a large-scale benchmark suite for open-world aerial object-goal search, with $3\times$ as many scenes and $18.7\times$ as many task instances as the largest existing benchmark for this task.
    \item AerialDojo-200K provides a high-quality dataset with extensive manual annotations. Twelve annotators spent two months manually annotating 109 landmarks, 2,099 objects, and 2,099 object anchors. Each task instance includes a collision-free reference trajectory and corresponding multi-view video recordings.
    \item We develop a unified evaluation framework for this task and evaluate 5 open-source and 4 closed-source MLLMs. The results reveal that there is still a long way to go toward achieving general-purpose aerial agents.
\end{itemize}

\section{Related Works}
We review related work on aerial object-goal search and aerial navigation benchmarks. We first discuss how existing methods combine spatial representations and semantic reasoning for autonomous search, then examine benchmark task formulations, dataset scale, and environmental coverage to contextualize AerialDojo-200K.

\input{tables/table1}

\subsection{Aerial Object-Goal Search} 
Aerial object-goal search requires an agent to reach a target object specified by a semantic description or reference image based solely on egocentric visual observations. Existing research can be grouped into two categories. On the one hand, existing aerial agents utilize spatio-semantic representations, such as semantic maps~\citep {zhou2026octmem}, scene graphs~\citep {gai2026ussnav}, and value maps~\citep{zhang2026apex}, to organize memories and plan over explored regions, frontiers, and object candidates. On the other hand, recent studies demonstrate that multimodal large models can leverage high-level semantic reasoning to interpret semantic goals~\citep{cityavos}, assess object candidates~\citep{pardyl2025flysearch}, and guide search decisions~\citep{traveluav, chen2026airhunt}. AerialOGS has broad potential applications~\citep{kim2025raven} in search and rescue \citep{feng2026u2udata+}, infrastructure inspection \citep{feng2024multi}, and autonomous delivery \citep{feng2024u2udata}, where operators may specify the object without knowing its exact location or providing a complete flight route. However, limited benchmark scale and heterogeneous evaluation protocols hinder systematic assessment of these methods across diverse environments. AerialDojo-200K addresses these limitations through large-scale, carefully annotated search tasks and a unified framework for evaluating performance and generalization.

\subsection{Aerial Benchmark}
Aerial tasks primarily include vision-and-language navigation (VLN) and object-goal navigation (ObjNav), with aerial object-goal search (OGS) representing a more challenging setting within ObjNav that emphasizes autonomous exploration and target identification. As shown in Table~\ref{tab1}, most existing aerial benchmarks focus on VLN tasks, where agents follow language instructions to reach specified locations. AerialVLN \citep{aerialvln} and CityNav \citep{citynav} provide 25.3K and 32.6K instruction-guided navigation tasks across 25 and 34 urban scenes, respectively. OpenUAV \citep{traveluav} extends this setting to a 6-DoF action space, providing 12.1K tasks across 22 urban and natural scenes, while OpenFly \citep{openfly} expands the task count to 103K across 18 scenes. Beyond instruction-guided navigation, UAV-ON \citep{uavon} supports instance-level ObjNav tasks through semantic goal specifications, with 11K tasks across 14 scenes. CityAVOS \citep{cityavos} further investigates OGS tasks using semantic and visual goal information, providing 2.4K tasks across six urban scenes. Despite these advances, existing benchmarks remain limited in both task scale and environmental coverage. Existing research relies on small, environment-specific benchmarks with heterogeneous action spaces and data formats, hindering large-scale training and cross-benchmark evaluation and limiting the scalability and generalizability of aerial agents.

\section{AerialDojo-200K}
\label{sec:environments}
\label{sec:benchmark}
\begin{figure}[h]
\centering
\includegraphics[width=\linewidth]{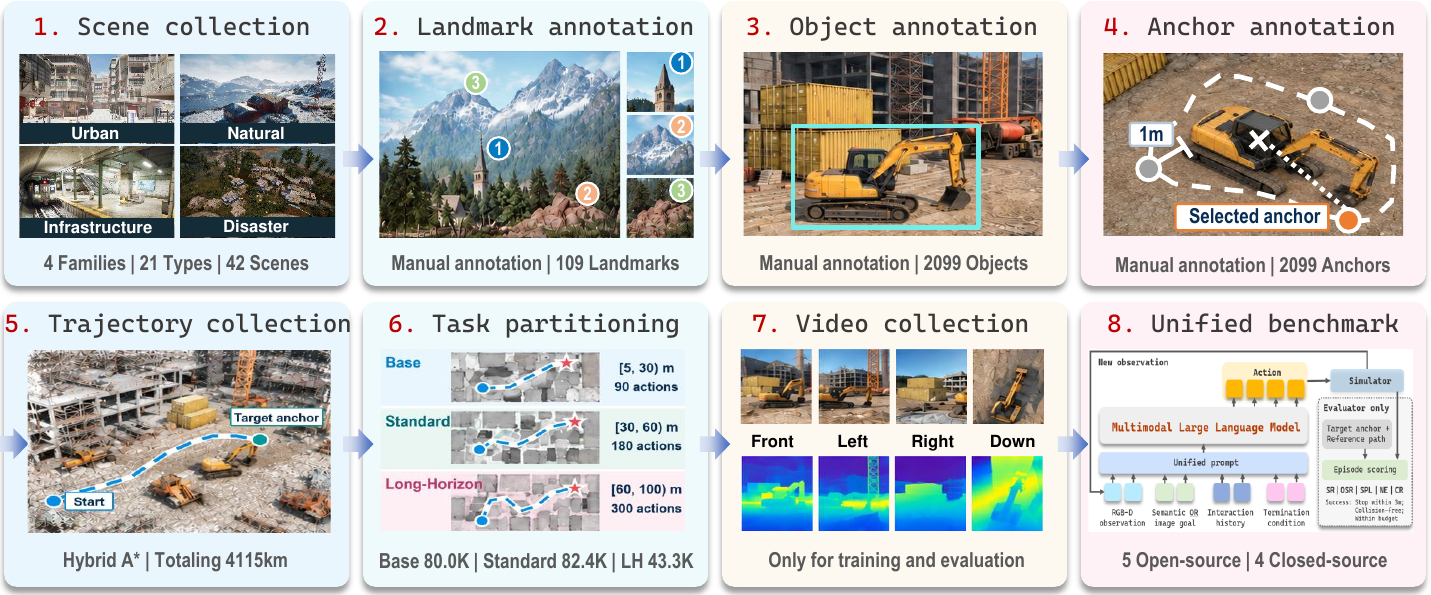}
\caption{Standard operating procedure for constructing AerialDojo-200K.}
\label{fig3}
\end{figure}

AerialDojo-200K, the first large-scale benchmark suite for open-world Aerial Object-Goal Search (AerialOGS), comprises 42 simulated scenes, 205K task instances, 4115 km of flight trajectories, and a unified evaluation framework. The construction process is shown in Figure \ref{fig3}.

\subsection{Simulator}
We build the AerialDojo-200K benchmark using Unreal Engine~\citep{unrealengine} and ProjectAirSim~\citep{projectairsim}, enabling AerialOGS across diverse, high-fidelity simulated scenes.

\begin{figure}[t]
\centering
\begin{subfigure}[t]{0.32\linewidth}
    \centering
    \includegraphics[width=\linewidth,height=2.5cm]{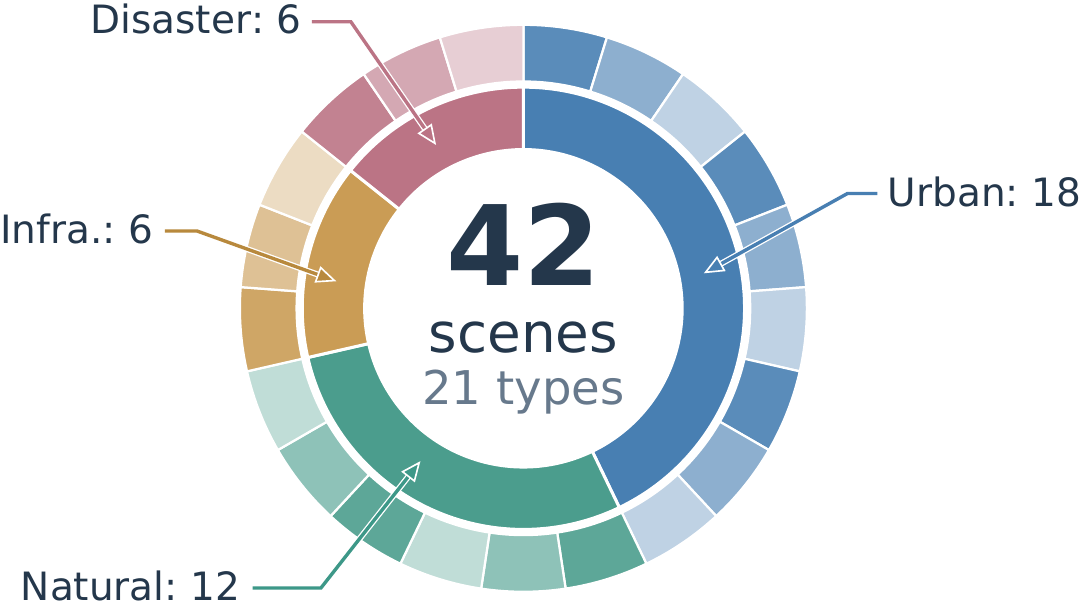}
    \caption{Simulator statistics}
    \label{fig:simulator_statistics}
\end{subfigure}
\hfill
\begin{subfigure}[t]{0.32\linewidth}
    \centering
    \includegraphics[width=\linewidth, height=2.5cm]{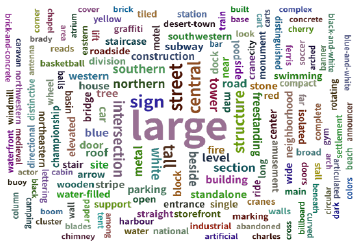}
    \caption{Object-name word cloud}
    \label{fig:object_wordcloud}
\end{subfigure}
\hfill
\begin{subfigure}[t]{0.32\linewidth}
    \centering
    \includegraphics[width=\linewidth, height=2.5cm]{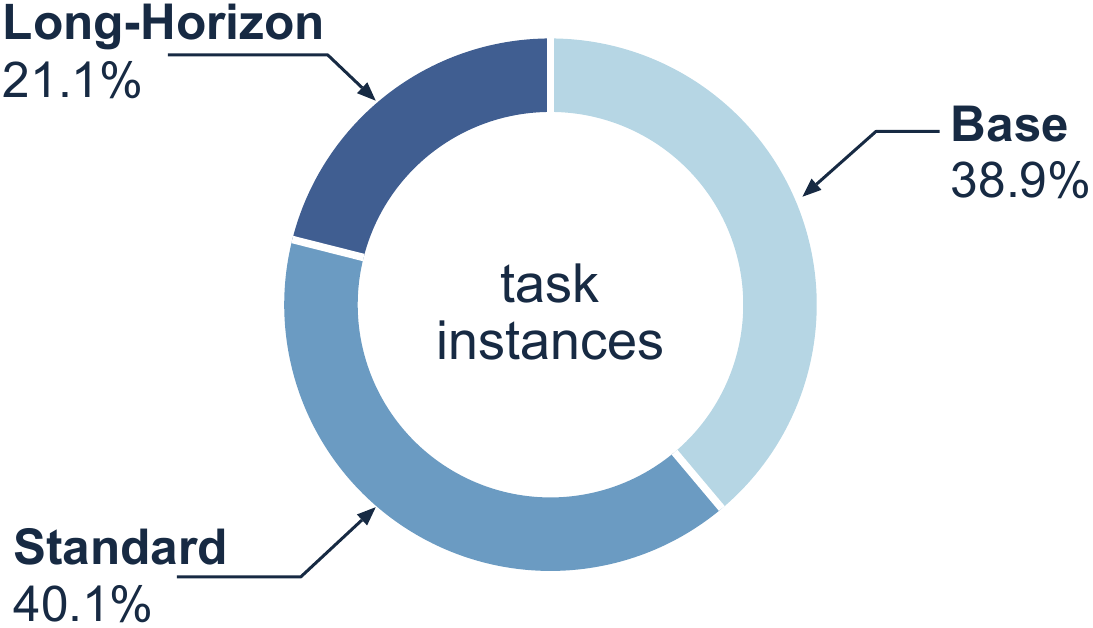}
    \caption{Task statistics}
    \label{fig:task_statistics}
\end{subfigure}
\caption{Simulator and dataset statistics of AerialDojo-200K.}
\label{fig2}
\vspace{-0.4cm}
\end{figure}

\textbf{Large-scale and comprehensive environment collection.}
AerialDojo-200K contains 42 scenes spanning four scene families and 21 scene types, with two instances per type. As shown in Table~\ref{tab1}, it provides three times as many scenes as UAV-ON~\citep{uavon}, the largest existing AerialOGS benchmark. As shown in Figure \ref{fig2}, the collection includes both open landscapes and constrained built spaces, with diverse layouts and obstacle configurations. Specifically, the urban family comprises malls, stadiums, amusement parks, parking lots, communities, neighborhoods, alleys, factories, and construction sites; the natural family covers deserts, forests, mountains, snowfields, islands, and coasts; the infrastructure family includes bridges, rail corridors, and harbors; and the disaster family depicts earthquake, flood, and explosion scenes. By incorporating infrastructure and disaster scenarios alongside urban and natural environments, AerialDojo-200K covers a broad range of settings relevant to real-world aerial deployment and applications. Together, these scenes form a comprehensive library of simulation environments for AerialOGS.

\textbf{Sensors.} AerialDojo-200K can collect synchronized RGB-D streams from four onboard cameras facing forward, left, right, and downward. Both RGB and depth images have a resolution of $640\times640$ pixels, and each camera has a $90^{\circ}$ field of view (FOV). All four views are sampled synchronously at 15 frames per second (FPS). At acquisition time $t$, the visual observation is represented as $o_t=\{(I_t^v,D_t^v)\mid v\in\mathcal{V}\}$ and $\mathcal{V}=\{\mathrm{front}, \mathrm{left}, \mathrm{right}, \mathrm{down}\}$, where $I_t^v$ and $D_t^v$ denote the RGB image and depth map from view $v$, respectively. The observation interface excludes GPS measurements, external localization signals, and global maps. Simulator ground truth, including object poses, semantic maps, and scene geometry, is also withheld from the agent. Consequently, the agent must interpret the goal, infer the surrounding spatial structure, and identify the target by integrating egocentric visual evidence over time.

\textbf{Actions.} AerialDojo-200K provides eight action types with discrete motion magnitudes, supporting four-DoF control of the aerial agent's position and yaw. Specifically, \texttt{MoveForward}, \texttt{MoveLeft}, and \texttt{MoveRight} allow horizontal translations of $1$, $3$, or $5$\,m, while \texttt{MoveUp} and \texttt{MoveDown} allow vertical translations of $1$ or $2$\,m. The rotational actions, \texttt{TurnLeft} and \texttt{TurnRight}, change the yaw angle by $15^{\circ}$ in the corresponding direction. The \texttt{Stop} action declares target discovery and terminates the episode. Together, these eight action types yield 16 discrete action choices, allowing the aerial agent to select movement magnitudes according to local geometry and target proximity. Backward translation is excluded to encourage forward-looking exploration and reduce redundant oscillatory movements. The aerial agent can reposition by combining yaw rotations with forward or lateral translations, enabling flexible exploration while maintaining a compact action interface.

\input{tables/table2}

\textbf{Aerial embodiment.} The aerial agent has a default wheelbase of $250\,\mathrm{mm}$ and a collision radius of $250\,\mathrm{mm}$. These settings are intended to represent compact commercial aerial platforms with wheelbases below $500\,\mathrm{mm}$. Both parameters are configurable.

\textbf{Manual annotation.} To ensure simulator quality, as shown in Table \ref{tab2}, 12 annotators spent two months manually annotating 109 landmarks, 2099 target objects, and 2099 object anchors across these scenes. AerialDojo-200K selects landmarks and target objects according to their roles in AerialOGS tasks. Landmarks serve as visually salient, readily identifiable spatial references, reflecting a human search strategy in unfamiliar environments: first locating a relevant landmark and then refining the target's position. Landmark selection is inherently scene-dependent; shopping malls or historic monuments may serve as urban landmarks, whereas distinctive mountain peaks or rock formations may provide natural landmarks. This variation motivates manual selection based on scene-level salience and contextual relevance. Target objects are selected for their practical search value and semantic distinguishability. Their descriptions must allow human annotators to unambiguously identify the intended instance, reducing ambiguous supervision during training. AerialDojo-200K uses object anchors for task evaluation because a centroid-based success region may lie entirely inside a large object, making successful arrival physically infeasible. An anchor is a collision-free viewpoint located $1$\,m outside the object's boundary and oriented toward its centroid. A vision-language model shortlists three candidate anchors based on visual informativeness and image--name correspondence. Human annotators then select one anchor per object as the spatial reference for task construction and success evaluation. We also have developed and open-sourced a lightweight annotation tool that standardizes the entire manual annotation workflow.

\subsection{Task and Dataset}
\label{sec:definition}
\textbf{Task definition.} AerialDojo-200K defines open-world AerialOGS tasks as $\mathcal{T}=\langle S, G\rangle$, where $S$ is the start pose and $G$ is the instance-level goal. $S=[x,y,z,\psi]$, where $(x,y,z)\in\mathbb{R}^{3}$ denotes its position and $\psi$ denotes its yaw angle. Unlike UAV-ON~\citep{uavon}, which randomly initializes the agent's start pose, we initialize each task instance at the anchor position of a selected object, with the yaw rotated by $180^\circ$ relative to the anchor orientation. This design can substantially reduce the risk of task failure due to collisions at initialization while better reflecting practical aerial search scenarios, in which an agent begins a new search from the location of a previously reached target. The goal is specified as $G=[\text{name}, \text{landmark}, \text{direction}, \text{representation}]$, comprising the target object's name, the name of its nearest landmark, its direction relative to that landmark, and a target representation provided as either a semantic description or a reference image. The direction takes one of eight compass orientations: north, south, east, west, northeast, northwest, southeast, and southwest. This goal reflects a common human search strategy in unfamiliar environments: using a recognizable landmark and a coarse direction to orient the search, then relying on visual evidence to identify the target object. AerialDojo-200K defines the task as \textbf{SemanticOGS} when the representation is a semantic description and \textbf{ImageOGS} when it is a reference image.

\textbf{Task execution.} An episode is specified by
\begin{equation}
\mathcal{X}=\langle \mathcal{T},\mathcal{E},\mathcal{B},\mathcal{O},\mathcal{A},\mathcal{C}\rangle,
\end{equation}
where $\mathcal{E}$ is the scene, $\mathcal{B}$ is the aerial embodiment, $\mathcal{O}$ and $\mathcal{A}$ are the observation and action spaces described above, and $\mathcal{C}$ specifies the termination conditions. At each decision step, the agent selects $a_t\in\mathcal{A}$ based on $G$ and its observation-action history. An episode terminates upon \texttt{Stop}, collision, or exhaustion of its action budget. Success requires an explicit \texttt{Stop} within $3$\,m of the target anchor, without any collision and within the assigned budget.

\textbf{Collection trajectory.} Candidate tasks pair distinct annotated objects within the same scene. We adapt Hybrid A*~\citep{dolgov2008practical} to the aerial agent's four-DoF action space to plan a collision-free reference trajectory from the start pose to the target anchor. Collision checking accounts for the agent's configured collision radius, and candidate pairs are discarded if no feasible trajectory is found. Let $\tau^{\mathrm{ref}}=(q_0,\ldots,q_K)$, where $q_k=(\mathbf{p}_k,\psi_k)$, $q_0=S$, $\mathbf{p}_0$ is the start position, and $\mathbf{p}_K=\mathbf{p}_g$ is the target anchor position. The translational trajectory length is
\begin{equation}
L(\tau^{\mathrm{ref}})
=\sum_{k=0}^{K-1}\|\mathbf{p}_{k+1}-\mathbf{p}_k\|_2.
\label{eq:reference_length}
\end{equation}
Pure yaw rotations contribute no translational length. Reference trajectories reach the target anchor itself, whereas evaluation accepts an explicit \texttt{Stop} within $3$\,m of the anchor. Target coordinates and reference trajectories are withheld from the agent during evaluation.

\textbf{Task partitioning.} We categorize tasks by reference trajectory length and use a detour ratio to filter out near-straight-line routes. Let $\mathbf{p}_0$ and $\mathbf{p}_g$ denote the start and target anchor positions, respectively. The Euclidean separation and detour ratio are defined as
\begin{equation}
d_E=\|\mathbf{p}_0-\mathbf{p}_g\|_2,
\qquad
\kappa=\frac{L(\tau^{\mathrm{ref}})}{d_E}.
\label{eq:task_complexity}
\end{equation}

\input{tables/table3}

We retain candidates satisfying $d_E>3$\,m, $5\,\mathrm{m}\leq L(\tau^{\mathrm{ref}})<100\,\mathrm{m}$, and $\kappa\geq1.05$. The first condition excludes starts already within the success region. The trajectory-length range balances meaningful search difficulty with a manageable evaluation horizon. The detour-ratio threshold filters out near-straight-line routes, aiming to reduce trivial tasks on which models may succeed uniformly and thereby improve the ability of success rate to distinguish model performance. Retained tasks are partitioned into Base, Standard, and Long-Horizon settings according to their reference trajectory lengths, with corresponding action budgets listed in Table~\ref{tab:task_settings}.

\textbf{Collection video.} We acquire synchronized multi-view RGB-D observations along reference trajectories using the sensor configuration described above. Quality checks flag missing camera outputs and abnormal image frames. Recordings are indexed by navigation task, camera view, and reference trajectory steps. The goal reference image used by ImageOGS is stored separately.

\textbf{Dataset statistics.} As shown in Figure \ref{fig2}, AerialDojo-200K contains 205,732 task instances, comprising 102,866 SemanticOGS and 102,866 ImageOGS instances across 42 scenes. The two instances of each navigation task share the same scene, start pose, target object, and reference trajectory, differing only in goal representation. The dataset covers 2,099 target objects across 254 semantic categories and reports a total planning distance of $4115.313$\,km.

\subsection{Benchmark}
\label{sec:benchmark_eval}

\textbf{Benchmark protocol.} To evaluate AerialOGS capabilities in open-world environments, AerialDojo-200K divides its 42 scenes into 21 in-distribution (ID) scenes and 21 out-of-distribution (OOD) scenes, with one instance of each scene type in each split. Within ID scenes, 80\% of the tasks are used for training and the remaining 20\% for in-distribution performance evaluation. Within OOD scenes, 20\% of the tasks form an adaptation subset, while the remaining 80\% are reserved for testing. Models trained on ID scenes are transferred to OOD scenes to evaluate generalization across distinct map layouts, obstacle configurations, and target instances within the same scene types. Inspired by adaptation with limited target-domain data, we allow models to use the designated 20\% OOD subset for scene-specific adaptation before evaluation on the held-out OOD test tasks. Evaluation covers SemanticOGS and ImageOGS across the Base, Standard, and Long-Horizon settings, using the action budgets in Table~\ref{tab:task_settings}.

\begin{wrapfigure}{r}{0.7\columnwidth}
\centering
\includegraphics[width=\linewidth]{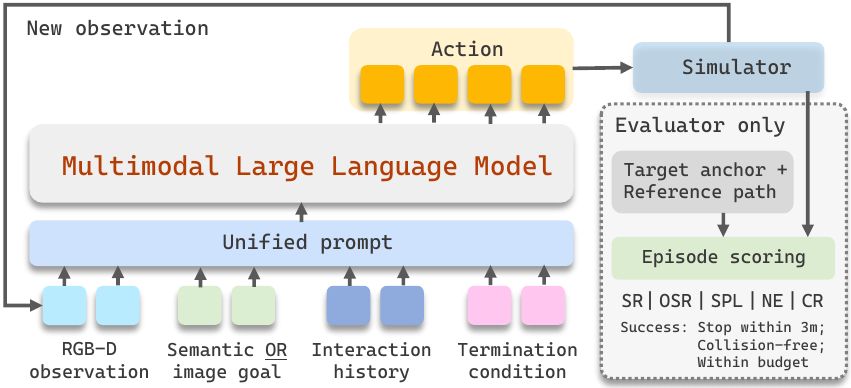}
\caption{The unified benchmark framework of AerialDojo-200K.}
\label{fig4}
\end{wrapfigure}
\textbf{Benchmark framework.} Figure~\ref{fig4} illustrates the closed-loop evaluation framework. Each episode resets the simulator to the task's scene and start pose. At each step, a model-specific adapter packages the current RGB-D observations, goal specification, termination conditions, and interaction history using the fixed prompt provided in Appendix~\ref{app:prompt}. The returns one action, which maps to the simulator's action interface using a common motion-magnitude configuration. Execution produces the next observation and updates the episode log. Episodes terminate upon \texttt{Stop}, collision, or budget exhaustion; infrastructure failures are logged and rerun separately. Target coordinates and reference trajectories remain accessible only to the evaluator.

\textbf{Benchmark metrics.} We evaluate navigation performance using five complementary metrics. Success rate (SR) measures the fraction of episodes in which the agent explicitly stops within $3$\,m of the target anchor without collision and within the action budget. Oracle success rate (OSR) measures whether the agent ever enters this region, regardless of its final stopping decision or subsequent failure. Distance to success (DTS) is the mean final Euclidean distance to the target anchor, measured in meters without subtracting the success radius. Success weighted by path length (SPL) evaluates navigation efficiency:
\begin{equation}
\mathrm{SPL}=\frac{1}{N}\sum_{i=1}^{N}
\sigma_i\frac{L_i}{\max(L_i,\ell_i)},
\end{equation}
where $N$ is the number of evaluated episodes, $\sigma_i$ indicates episode success, $L_i=L(\tau_i^{\mathrm{ref}})$ is the reference trajectory length and $\ell_i$ is the executed trajectory length. This formulation uses the planned reference for normalization. Collision rate (CR) is the fraction of episodes containing a collision. All metrics are computed over both successful and failed episodes.

\section{Experiments}
We conduct experiments on AerialDojo-200K to evaluate existing MLLMs on AerialOGS. We first introduce the baseline models and experimental setup, followed by quantitative comparisons across scene families and task setting. We further analyze how the success threshold affects model performance and rankings.

\subsection{Baselines and Setup}
\label{sec:exp_baselines}
\textbf{Baselines.} We evaluate five open-source MLLMs, Qwen3-VL-4B, InternVL3.5-8B, Ministral 3 8B, Phi-4-multimodal, and MiniCPM-V 4.6, and four closed-source MLLMs, GPT-5.6 Sol, Grok-4.5, Gemini 3.8 Flash, and Claude Opus 5. We test these models under the Base, Standard, and Long-Horizon settings. For brevity, the main tables report model-setting combinations with nonzero overall success rates, covering all nine MLLMs on Base tasks, GPT-5.6 Sol and Claude Opus 5 on Standard tasks.

\textbf{Evaluation setup.} Base and Standard tasks have reference lengths of $[5,30)$\,m and $[30,60)$\,m and action budgets of 90 and 180, respectively. We average the all map-level scores equally within each family, then average the four family means equally for overall results. Rates are percentages, and DTS is in meters. The prompt specifies stopping within 3\,m, our primary success radius; 5\,m is a supplementary evaluation threshold.

\input{tables/table4}
\input{tables/table5}

\subsection{Evaluation Results}
\label{sec:exp_results}
\textbf{Performance on base and standard tasks.} Tables~\ref{tab:exp_base_overall}--\ref{tab:exp_standard} show substantial performance differences among the evaluated models, alongside consistently low task completion rates. At the primary 3\,m threshold, Claude Opus 5 achieves the highest Base SR of 3.20\%, followed by GPT-5.6 Sol (2.81\%) and Gemini 3.8 Flash (2.40\%). Qwen3-VL-4B leads the open-source models at 0.71\%, exceeding Grok-4.5 at 0.25\%, indicating that the advantage of closed-source models is not uniform. Performance deteriorates markedly on Standard tasks: GPT-5.6 Sol and Claude Opus 5 achieve SRs of only 0.26\% and 0.49\%, respectively. Their OSRs also decrease to 0.26\% and 1.32\%, showing that even reaching the target region is uncommon. Together, these results highlight the difficulty of maintaining successful search beyond the Base setting.

\textbf{Success-radius sensitivity.}
Relaxing the success radius from 3\,m to 5\,m substantially changes both success rates and model rankings. On Base tasks, Gemini 3.8 Flash increases from 2.40\% to 11.16\% SR, an improvement of 8.76 percentage points, overtaking GPT-5.6 Sol (8.16\%) and Claude Opus 5 (7.38\%). A ranking reversal also occurs on Standard tasks: Claude Opus 5 leads at 3\,m, whereas GPT-5.6 Sol leads at 5\,m with 2.30\% SR versus 1.29\%. These changes demonstrate that model comparisons depend strongly on the spatial tolerance used to define success. The higher scores at 5\,m reflect a more permissive evaluation criterion and do not establish improved navigation behavior. We therefore report both thresholds while retaining the prompt-aligned 3\,m criterion as the primary measure of precise task completion.

\begin{figure*}[t]
\centering
\includegraphics[width=\textwidth]{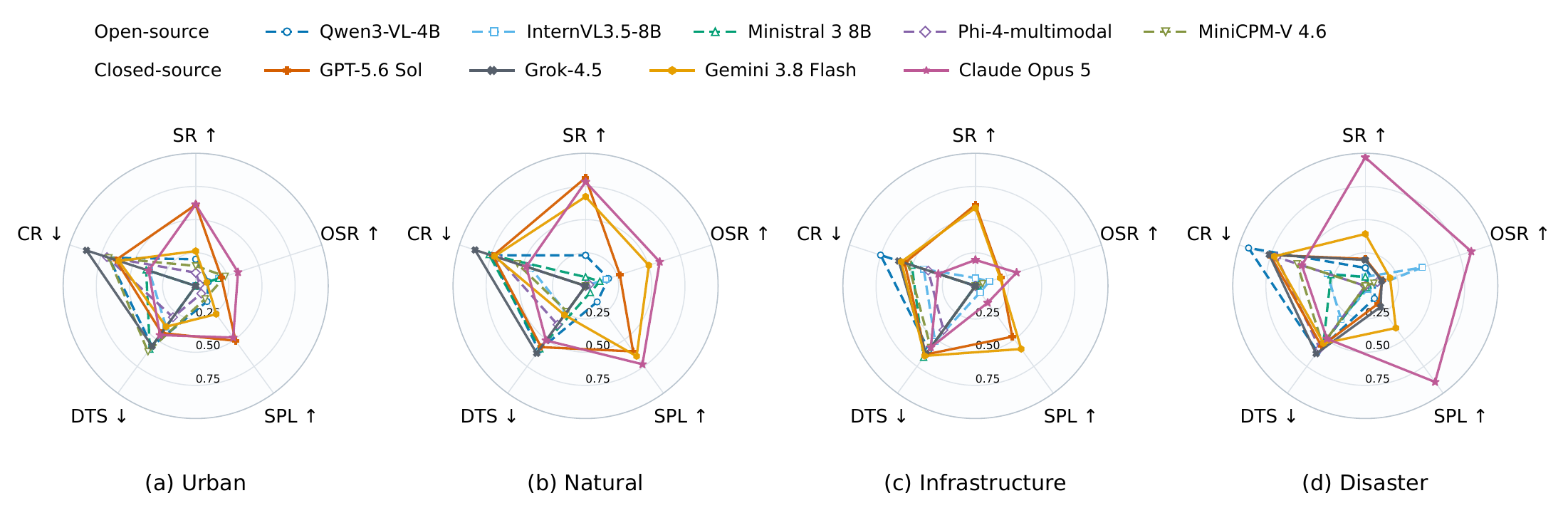}
\caption{Base-task performance across four scene families at the 3\,m success threshold.
Each curve represents one model. From the center to the outer edge, SR and SPL range from 0 to 5\%, OSR from 0 to 15\%, DTS from 40 to 0\,m, and CR from 100 to 0\%; outward therefore indicates better performance. Ring labels show normalized radial values.}
\label{fig5}
\end{figure*}

\textbf{Performance differences across scene families.}
Figure~\ref{fig5} reveals scene-dependent model strengths that are obscured by overall averages. At 3\,m, GPT-5.6 Sol achieves the highest SR on Natural and Infrastructure scenes, at 4.08\% and 3.06\%, respectively. Claude Opus 5 leads on Disaster scenes with 4.85\% SR, but achieves only 0.98\% on Infrastructure scenes. On Urban scenes, Claude Opus 5 and GPT-5.6 Sol obtain nearly identical SRs of 3.07\% and 3.06\%. The radar profiles further show that strengths in successful completion do not consistently coincide with strengths in collision avoidance or final target distance. Thus, no evaluated model dominates all scene families and metrics, emphasizing the need for family-wise evaluation to expose differences in capability across the evaluated environments.

\textbf{A long way to generalist aerial agents.}
Collectively, these results reveal a substantial gap between current MLLMs under the evaluated framework and reliable generalist aerial agents. The best reported SR remains only 3.20\% on Base tasks and 0.49\% on Standard tasks at the primary threshold. Moreover, reaching the target region does not ensure successful completion: Claude Opus 5 achieves a Base OSR of 7.84\% but an SR of only 3.20\%, alongside a collision rate of 58.96\%. This gap captures failures to convert target-region visits into valid completion, although the aggregate metrics cannot identify their individual causes. Low completion rates, frequent collisions, and uneven performance across scene families collectively indicate that integrating effective exploration, precise target localization, appropriate stopping, and collision avoidance remains a major challenge on the path toward generalist aerial agents.

\section{Conclusion}
We present AerialDojo-200K, a large-scale benchmark suite for open-world aerial object-goal search. It comprises 205,732 semantic-goal and image-goal task instances across 42 simulated scenes from four scene families, with three task settings and a unified observation, action, and evaluation interface. Our evaluation of existing MLLMs reveals substantial limitations: the best reported success rates at the primary 3\,m threshold reach only 3.20\% on Base tasks and 0.49\% on Standard tasks. Together with frequent collisions and gaps between target-region visits and successful completion, these findings show that current MLLMs under the evaluated framework remain far from reliable generalist aerial agents. AerialDojo-200K provides a common testbed for developing and evaluating methods that integrate exploration, instance-level target grounding, precise stopping, and collision avoidance.

\bibliography{AerialDojo}
\bibliographystyle{AerialDojo}

\clearpage
\appendix
\input{appendix/A}
\input{appendix/C}
\input{appendix/D}

\end{document}

%% file: math_commands.tex
\usepackage{amsmath,amsfonts,bm}

\def\eqref#1{equation~\ref{#1}}

\def\1{\bm{1}}

\DeclareMathAlphabet{\mathsfit}{\encodingdefault}{\sfdefault}{m}{sl}
\SetMathAlphabet{\mathsfit}{bold}{\encodingdefault}{\sfdefault}{bx}{n}



%% file: tables/table1.tex
\begin{table*}[t]
\centering
\caption{Comparison of aerial benchmarks. $N_{\mathrm{task}}$: the number of task instances; $F_{\mathrm{scene}}$: Scene families; $N_{\mathrm{scene}}$: the number of scenes. VLN: vision-and-language navigation; ObjNav: object-goal navigation; OGS: object-goal search. DoF: degrees of fredom. U: urban; N: natural; I: infrastructure; D: disaster.}
\label{tab1}
\begingroup
\fontsize{8.5}{10.2}\selectfont
\setlength{\tabcolsep}{2.5pt}
\renewcommand{\arraystretch}{1.22}
\begin{tabularx}{\textwidth}{@{}
>{\raggedright\arraybackslash}p{.16\textwidth}
>{\centering\arraybackslash}p{.05\textwidth}
>{\raggedright\arraybackslash}p{.07\textwidth}
>{\centering\arraybackslash}p{.07\textwidth}
>{\centering\arraybackslash}p{.11\textwidth}
>{\raggedright\arraybackslash}X
>{\centering\arraybackslash}p{.08\textwidth}
>{\centering\arraybackslash}p{.085\textwidth}
>{\centering\arraybackslash}p{.065\textwidth}@{}}
\toprule
\textbf{Benchmark} & \textbf{Year} & \textbf{Task} & \textbf{Action} & \textbf{Goal Type} & \textbf{Goal Specification} & $N_{\mathrm{task}}$ & $F_{\mathrm{scene}}$ & $N_{\mathrm{scene}}$ \\
\midrule
AerialVLN & 2023 & VLN & 4-DoF & Location & Movement instruction & 25.3K & U & 25 \\
CityNav & 2025 & VLN & 4-DoF & Location & Movement instruction & 32.6K & U & 34 \\
OpenUAV & 2025 & VLN & 6-DoF & Location & Movement instruction & 12.1K & U,N & 22 \\
OpenFly & 2026 & VLN & 4-DoF & Location & Movement instruction & 103K & U & 18 \\
\midrule
UAV-ON & 2025 & ObjNav & 4-DoF & Instance & Semantic & 11K & U,N & 14 \\
CityAVOS & 2026 & OGS & 4-DoF & Instance & Semantic and Image & 2.4K & U & 6 \\
\midrule
\rowcolor{lastrowblue}
\textbf{AerialDojo-200K} & \textbf{2026} & \textbf{OGS} & \textbf{4-DoF} & \textbf{Instance} & \textbf{Semantic \textcolor{red}{or} image} & \textbf{\textcolor{red}{200K}} & \textbf{\textcolor{red}{U,N,I,D}} & \textbf{\textcolor{red}{42}} \\
\bottomrule
\end{tabularx}
\endgroup
\end{table*}

%% file: tables/table2.tex
\begin{table}[!t]
\centering
\small
\fontsize{8.5}{10.2}\selectfont
\setlength{\tabcolsep}{3pt}
\caption{Dataset statistics. LM: landmarks; Base, Standard, and LH denote task-instance counts for the Base, Standard, and Long-Horizon setting, respectively, including both semantic-goal and image-goal instances; ALL is their sum. Path reports the cumulative planned path length in kilometers, counted once per underlying navigation task.}
\label{tab2}
\resizebox{\linewidth}{!}{%
\begin{tabular}{@{}lrrrrrrrrrr@{}}
\toprule
Family & Types & Scenes & LM & Objects & Anchors & Base & Standard & LH & ALL & Path (km) \\
\midrule
Urban & 9 & 18 & 57 & 1,030 & 1,030 & 42,232 & 55,840 & 26,992 & 125,064 & 2,607.253 \\
Natural & 6 & 12 & 21 & 569 & 569 & 21,668 & 15,188 & 12,976 & 49,832 & 1,008.083 \\
Infrastructure & 3 & 6 & 20 & 280 & 280 & 8,934 & 7,800 & 3,080 & 19,814 & 357.067 \\
Disaster & 3 & 6 & 11 & 220 & 220 & 7,140 & 3,596 & 286 & 11,022 & 142.910 \\
\midrule
\rowcolor{lastrowblue}
\textbf{Total} & \textbf{21} & \textbf{42} & \textbf{109} & \textbf{2,099} & \textbf{2,099} & \textbf{79,974} & \textbf{82,424} & \textbf{43,334} & \textbf{205,732} & \textbf{4,115.313} \\
\bottomrule
\end{tabular}%
}
\end{table}

%% file: tables/table3.tex
\begin{wraptable}{r}{0.4\columnwidth}
\centering
\caption{Task settings.}
\label{tab:task_settings}
\small
\renewcommand{\arraystretch}{1.15}
\begin{tabular*}{\linewidth}{@{\extracolsep{\fill}}lcc@{}}
\toprule
Setting & $L(\tau^{\mathrm{ref}})$ & \shortstack{Action\\budget} \\
\midrule
Base & $[5,30)$ & 90 \\
Standard & $[30,60)$ & 180 \\
Long-Horizon & $[60,100)$ & 300 \\
\bottomrule
\end{tabular*}
\end{wraptable}

%% file: tables/table4.tex
\begin{table}[!h]
\centering
\caption{Base-task performance across four scene families. SR, OSR, SPL, and CR are percentages; DTS is in meters. Arrows indicate the preferred direction; bold denotes the best available value.}
\label{tab:exp_base_overall}
\begingroup
\definecolor{bestcell}{HTML}{ECF4FF}
\newcommand{\metricvar}[2]{\mbox{#1{\tiny$\pm$#2}}}
\newcommand{\bestmetric}[2]{\cellcolor{bestcell}\metricvar{\textbf{#1}}{#2}}
\fontsize{8}{10}\selectfont
\setlength{\tabcolsep}{1.5pt}
\renewcommand{\arraystretch}{1.12}
\begin{tabular*}{\linewidth}{@{\extracolsep{\fill}}lrrrrrrr@{}}
\toprule
\rowcolor{expblue}Model & \multicolumn{5}{c}{3\,m (primary)} & \multicolumn{2}{c}{5\,m} \\
\cmidrule(lr){2-6}\cmidrule(lr){7-8}
 & SR$\uparrow$ & OSR$\uparrow$ & SPL$\uparrow$ & DTS$\downarrow$ & CR$\downarrow$ & SR$\uparrow$ & OSR$\uparrow$ \\
\midrule
\rowcolor{expgroup}\multicolumn{8}{l}{\textit{Open-source models}} \\
Qwen3-VL-4B & \metricvar{0.71}{0.81} & \metricvar{1.50}{1.33} & \metricvar{0.51}{0.60} & \metricvar{16.24}{1.85} & \bestmetric{22.17}{14.32} & \metricvar{3.85}{2.92} & \metricvar{7.99}{3.93} \\
InternVL3.5-8B & \metricvar{0.16}{0.30} & \metricvar{3.41}{2.36} & \metricvar{0.11}{0.22} & \metricvar{25.42}{11.63} & \metricvar{62.06}{16.89} & \metricvar{1.54}{1.74} & \metricvar{12.04}{4.47} \\
Ministral 3 8B & \metricvar{0.17}{0.31} & \metricvar{1.28}{1.48} & \metricvar{0.09}{0.21} & \metricvar{16.23}{3.43} & \metricvar{51.63}{24.41} & \metricvar{1.64}{1.18} & \metricvar{11.37}{5.20} \\
Phi-4-multimodal & \metricvar{0.13}{0.35} & \metricvar{0.60}{1.01} & \metricvar{0.09}{0.25} & \metricvar{23.52}{7.47} & \metricvar{37.96}{16.67} & \metricvar{1.65}{1.79} & \metricvar{9.22}{6.17} \\
MiniCPM-V 4.6 & \metricvar{0.19}{0.53} & \metricvar{1.35}{2.41} & \metricvar{0.15}{0.43} & \metricvar{20.26}{6.44} & \metricvar{42.47}{11.07} & \metricvar{1.47}{1.98} & \metricvar{9.91}{3.84} \\
\addlinespace[2pt]
\rowcolor{expgroup}\multicolumn{8}{l}{\textit{Closed-source models}} \\
GPT-5.6 Sol & \metricvar{2.81}{2.42} & \metricvar{3.06}{2.44} & \metricvar{2.20}{1.87} & \metricvar{17.90}{4.01} & \metricvar{32.91}{14.45} & \metricvar{8.16}{3.78} & \metricvar{11.22}{4.23} \\
Grok-4.5 & \metricvar{0.25}{0.69} & \metricvar{0.49}{1.39} & \metricvar{0.25}{0.69} & \bestmetric{15.46}{1.81} & \metricvar{22.18}{12.44} & \metricvar{2.46}{2.28} & \metricvar{3.69}{2.65} \\
Gemini 3.8 Flash & \metricvar{2.40}{2.23} & \metricvar{3.68}{4.32} & \metricvar{2.37}{2.21} & \metricvar{21.61}{7.07} & \metricvar{33.95}{12.21} & \bestmetric{11.16}{8.27} & \metricvar{17.90}{8.25} \\
Claude Opus 5 & \bestmetric{3.20}{2.04} & \bestmetric{7.84}{4.60} & \bestmetric{2.83}{2.04} & \metricvar{19.69}{5.97} & \metricvar{58.96}{15.94} & \metricvar{7.38}{3.73} & \bestmetric{18.48}{5.64} \\
\bottomrule
\end{tabular*}
\endgroup
\end{table}

%% file: tables/table5.tex
\begin{table}[h]
\centering
\caption{Standard-task performance across four scene families. SR, OSR, SPL, and CR are percentages; DTS is in meters. Arrows is the preferred direction; bold denotes the best available value.}
\label{tab:exp_standard}
\begingroup
\definecolor{bestcell}{HTML}{ECF4FF}
\newcommand{\metricstd}[2]{\mbox{#1{\tiny$\pm$#2}}}
\newcommand{\bestmetric}[2]{\cellcolor{bestcell}\metricstd{\textbf{#1}}{#2}}
\fontsize{8}{10}\selectfont
\setlength{\tabcolsep}{1.5pt}
\renewcommand{\arraystretch}{1.12}
\begin{tabular*}{\linewidth}{@{\extracolsep{\fill}}lrrrrrrr@{}}
\toprule
\rowcolor{expblue}Model & \multicolumn{5}{c}{3\,m (primary)} & \multicolumn{2}{c}{5\,m} \\
\cmidrule(lr){2-6}\cmidrule(lr){7-8}
 & SR$\uparrow$ & OSR$\uparrow$ & SPL$\uparrow$ & DTS$\downarrow$ & CR$\downarrow$ & SR$\uparrow$ & OSR$\uparrow$ \\
\midrule
GPT-5.6 Sol & \metricstd{0.26}{0.72} & \metricstd{0.26}{0.72} & \metricstd{0.20}{0.57} & \metricstd{37.66}{5.47} & \bestmetric{36.57}{16.79} & \bestmetric{2.30}{2.02} & \metricstd{2.30}{2.02} \\
Claude Opus 5 & \bestmetric{0.49}{0.91} & \bestmetric{1.32}{2.33} & \bestmetric{0.40}{0.73} & \bestmetric{36.49}{8.89} & \metricstd{56.30}{21.31} & \metricstd{1.29}{1.91} & \bestmetric{2.61}{3.84} \\
\bottomrule
\end{tabular*}
\endgroup
\end{table}

%% file: appendix/A.tex
\clearpage
\section{Prompt}
\label{app:prompt}
\begin{figure}[htbp]
    \centering
    \includegraphics[width=\linewidth]{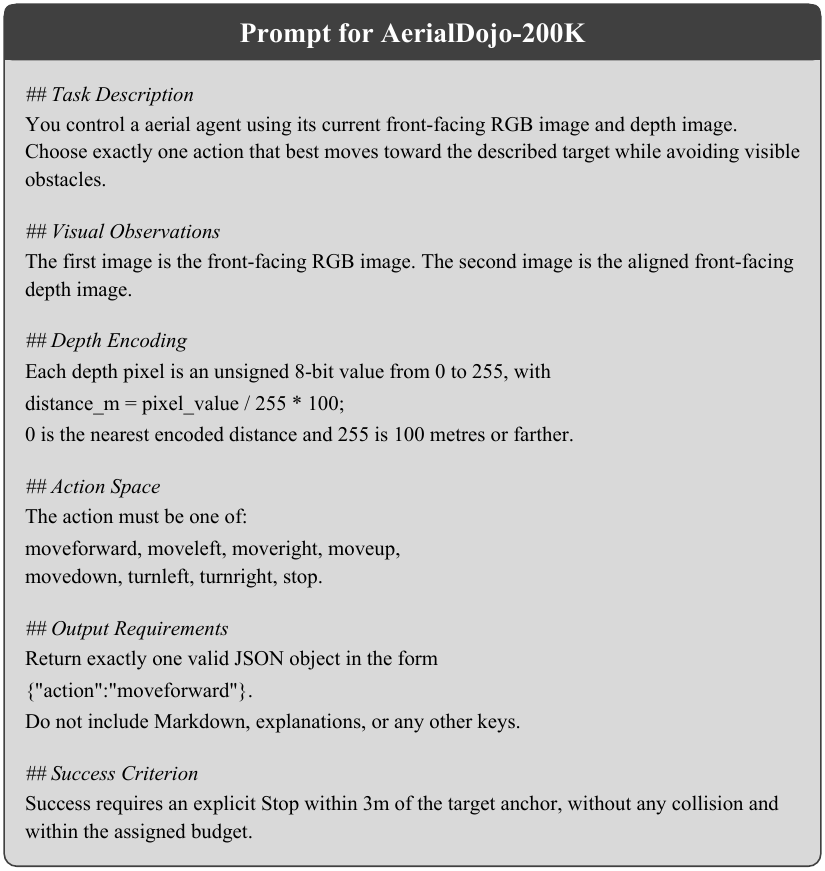}
    \caption{Prompt for MLLM evaluation in AerialDojo-200K.}
    \label{fig:prompt}
\end{figure}

Figure~\ref{fig:prompt} presents the evaluation prompt for SemanticOGS in AerialDojo-200K. The prompt specifies the front-facing RGB-D inputs, depth encoding, eight available actions, aerial embodiment configurations, termination Conditions, and the required single-action JSON response. It defines successful completion as an explicit \texttt{Stop} within 3\,m of the target anchor, without collisions and within the assigned action budget.

%% file: appendix/C.tex
\clearpage
\section{Task Composition and Reference Lengths}
\begin{figure}[!h]\centering
\includegraphics[width=\linewidth]{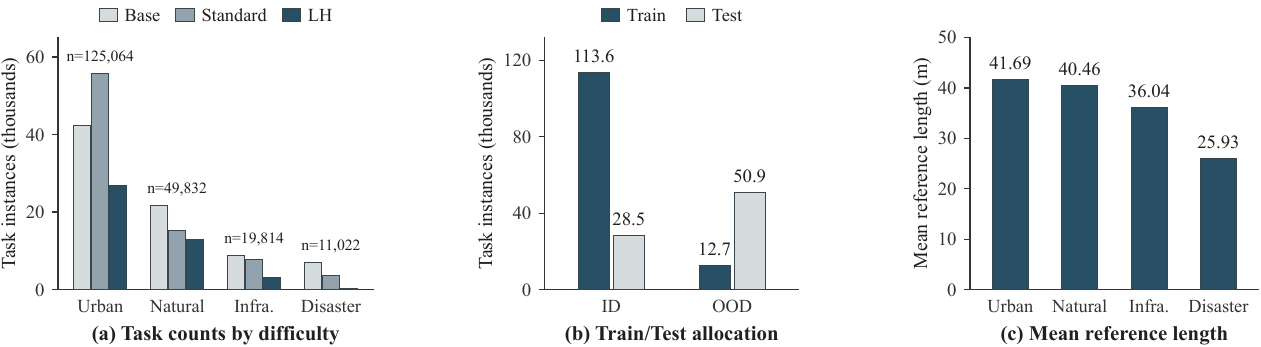}
\caption{Task composition and reference lengths. (a) Goal-conditioned instance counts by family and setting. (b) Train/Test counts in the ID and OOD scene groups. Counts in (a,b) include both goals. (c) Mean reference length, weighted by navigation-task count within each family.}\label{fig:taskcomposition}
\end{figure}

AerialDojo-200K contains 79,974 Base, 82,424 Standard, and 43,334 Long-Horizon task instances, representing 38.87\%, 40.06\%, and 21.06\% of the release, respectively. Figure~\ref{fig:taskcomposition} separates task volume, Train/Test allocation, and reference length so that a large task set is not confused with a task set containing longer routes.

Urban scenes contribute 60.79\% of all instances, followed by Natural (24.22\%), Infrastructure (9.63\%), and Disaster (5.36\%). Mall scenes alone contribute 28.82\%. Scene-level counts range from 190 in \texttt{N\_Mountain\_1} to 40,214 in \texttt{U\_Mall\_0}. These unequal counts motivate reporting the aggregation rule explicitly when comparing methods; a task-weighted score and an equally weighted scene-family score represent different evaluation populations.

\section{Scene Partitions and Target Coverage}
The scene partition assigns one instance of each type to ID and the other to OOD, giving 21 scenes per group. Equal scene counts do not imply equal task counts: ID scenes contain 142,094 instances (69.07\%), whereas OOD scenes contain 63,638 (30.93\%). Table~\ref{tab:splitdifficulty} reports the actual integer allocation. ID scenes contain 113,642 training and 28,452 test instances; the corresponding OOD counts are 12,712 and 50,926. Their training proportions are 79.98\% and 19.98\%, respectively. These realized proportions are close to the nominal 80/20 and 20/80 allocations.

\begin{table}[!htbp]\centering\small
\caption{Task-instance counts by scene group, local Train/Test subset, and difficulty. Both goal modalities are included. ID/OOD identifies the scene group; Train/Test identifies the task subset within that group.}\label{tab:splitdifficulty}
\begin{tabularx}{\linewidth}{>{\raggedright\arraybackslash}X>{\raggedright\arraybackslash}X*{4}{>{\raggedleft\arraybackslash}X}}
\toprule
Scenes & Task subset & Base & Standard & LH & ALL \\
\midrule
ID & Train & 40,640 & 48,398 & 24,604 & 113,642 \\
ID & Test & 10,240 & 12,060 & 6,152 & 28,452 \\
\midrule
OOD & Train & 5,764 & 4,452 & 2,496 & 12,712 \\
OOD & Test & 23,330 & 17,514 & 10,082 & 50,926 \\
\midrule
\rowcolor[HTML]{ECF4FF}\textbf{Total} & \textbf{All} & \textbf{79,974} & \textbf{82,424} & \textbf{43,334} & \textbf{205,732} \\
\bottomrule
\end{tabularx}
\end{table}

%% file: appendix/D.tex
\clearpage
\section{Inventory}
As summarized in Table~\ref{tab:inventory}, AerialDojo-200K contains
205,732 search task instances across 42 simulation scenes, covering
21 scene types in four scene families: urban, natural,
infrastructure, and disaster. Each scene type includes one
in-distribution (ID) and one out-of-distribution (OOD) instance.
Together, these scenes contain 109 landmarks, 2,099 annotated
objects, and 2,099 anchors. The dataset comprises 79,974 Base,
82,424 Standard, and 43,334 Long-Horizon task instances, with
both goal modalities included in these counts. The reference
trajectories span a total of 4,115.313~km, with each trajectory
counted once.

\input{tables/table10}

%% file: tables/table10.tex
\begingroup
\fontsize{9}{10}\selectfont
\setlength{\tabcolsep}{3pt}
\renewcommand{\arraystretch}{1.0}
\setlength{\LTleft}{0pt}
\setlength{\LTright}{0pt}
\setlength{\LTcapwidth}{\linewidth}
\begin{longtable}{@{\extracolsep{\fill}}lrrrrrrr@{}}
\caption{Scene inventory and search task statistics in AerialDojo-200K.
For each scene, LM reports the number of landmarks, and Obj./Anc.
reports the numbers of annotated objects and anchors, respectively.
Base, Standard, and LH report the numbers of search task instances in the Base, Standard, and Long-Horizon settings; ALL is their sum.
Task instance counts include both goal modalities.
Path (km) reports the total length of reference trajectories,
with each trajectory counted once.
Scene suffixes \texttt{\_0} and \texttt{\_1} indicate
in-distribution (ID) and out-of-distribution (OOD) scenes, respectively.}\label{tab:inventory}\\
\toprule Scene & LM & Obj./Anc. & Base & Standard & LH & ALL & Path (km) \\
\midrule
\endfirsthead
\multicolumn{8}{l}{\textit{Table~\ref{tab:inventory} continued}}\\
\toprule Scene & LM & Obj./Anc. & Base & Standard & LH & ALL & Path (km) \\
\midrule
\endhead
\midrule\multicolumn{8}{r}{\textit{Continued on the next page}}\\
\endfoot
\bottomrule
\endlastfoot
\rowcolor[HTML]{ECF4FF}\multicolumn{8}{@{}l}{\textbf{Urban}} \\
Mall\_0 & 6 & 123/123 & 12,412 & 19,652 & 8,150 & 40,214 & 839.245 \\
Mall\_1 & 5 & 102/102 & 5,390 & 7,144 & 6,536 & 19,070 & 459.569 \\
Stadium\_0 & 5 & 64/64 & 4,344 & 8,032 & 146 & 12,522 & 213.410 \\
Stadium\_1 & 1 & 41/41 & 1,234 & 2,010 & 1,156 & 4,400 & 97.323 \\
Park\_0 & 2 & 14/14 & 346 & 90 & 6 & 442 & 5.077 \\
Park\_1 & 2 & 46/46 & 1,452 & 1,616 & 600 & 3,668 & 70.006 \\
ParkingLot\_0 & 7 & 78/78 & 1,552 & 5,788 & 3,140 & 10,480 & 265.216 \\
ParkingLot\_1 & 1 & 26/26 & 676 & 198 & 0 & 874 & 8.540 \\
Community\_0 & 4 & 65/65 & 392 & 740 & 1,128 & 2,260 & 65.735 \\
Community\_1 & 3 & 31/31 & 300 & 384 & 104 & 788 & 14.582 \\
Neighborhood\_0 & 1 & 32/32 & 1,738 & 996 & 26 & 2,760 & 36.245 \\
Neighborhood\_1 & 2 & 63/63 & 4,486 & 680 & 32 & 5,198 & 51.460 \\
Alley\_0 & 3 & 27/27 & 1,188 & 714 & 100 & 2,002 & 26.514 \\
Alley\_1 & 2 & 67/67 & 2,506 & 1,586 & 306 & 4,398 & 65.324 \\
Factory\_0 & 7 & 80/80 & 1,052 & 1,642 & 1,048 & 3,742 & 86.301 \\
Factory\_1 & 2 & 50/50 & 512 & 324 & 124 & 960 & 16.148 \\
ConstructionSite\_0 & 3 & 59/59 & 1,738 & 3,462 & 3,640 & 8,840 & 232.868 \\
ConstructionSite\_1 & 1 & 62/62 & 914 & 782 & 750 & 2,446 & 53.690 \\
\rowcolor[HTML]{ECF4FF}\multicolumn{8}{@{}l}{\textbf{Natural}} \\
Desert\_0 & 3 & 64/64 & 3,010 & 5,008 & 6,294 & 14,312 & 382.722 \\
Desert\_1 & 1 & 29/29 & 700 & 236 & 28 & 964 & 11.950 \\
Forest\_0 & 1 & 54/54 & 7,220 & 412 & 0 & 7,632 & 64.708 \\
Forest\_1 & 1 & 28/28 & 1,012 & 626 & 84 & 1,722 & 24.887 \\
Mountain\_0 & 2 & 53/53 & 648 & 356 & 400 & 1,404 & 26.958 \\
Mountain\_1 & 1 & 32/32 & 132 & 54 & 4 & 190 & 2.262 \\
Snowfield\_0 & 1 & 21/21 & 478 & 214 & 2 & 694 & 8.153 \\
Snowfield\_1 & 1 & 44/44 & 1,706 & 812 & 1,036 & 3,554 & 72.382 \\
Island\_0 & 5 & 79/79 & 2,942 & 3,362 & 1,746 & 8,050 & 161.095 \\
Island\_1 & 2 & 60/60 & 1,098 & 922 & 814 & 2,834 & 60.992 \\
Coast\_0 & 1 & 50/50 & 1,590 & 2,110 & 2,082 & 5,782 & 141.467 \\
Coast\_1 & 2 & 55/55 & 1,132 & 1,076 & 486 & 2,694 & 50.507 \\
\rowcolor[HTML]{ECF4FF}\multicolumn{8}{@{}l}{\textbf{Infrastructure}} \\
Bridge\_0 & 3 & 28/28 & 382 & 334 & 208 & 924 & 17.401 \\
Bridge\_1 & 3 & 20/20 & 240 & 152 & 20 & 412 & 5.796 \\
RailCorridor\_0 & 3 & 50/50 & 2,326 & 2,986 & 666 & 5,978 & 110.070 \\
RailCorridor\_1 & 5 & 54/54 & 1,592 & 674 & 36 & 2,302 & 27.301 \\
Harbor\_0 & 4 & 110/110 & 4,336 & 3,466 & 1,974 & 9,776 & 184.783 \\
Harbor\_1 & 2 & 18/18 & 58 & 188 & 176 & 422 & 11.716 \\
\rowcolor[HTML]{ECF4FF}\multicolumn{8}{@{}l}{\textbf{Disaster}} \\
Earthquake\_0 & 2 & 50/50 & 1,490 & 40 & 0 & 1,530 & 10.934 \\
Earthquake\_1 & 3 & 49/49 & 1,132 & 668 & 212 & 2,012 & 32.151 \\
Flood\_0 & 3 & 25/25 & 944 & 610 & 0 & 1,554 & 20.214 \\
Flood\_1 & 1 & 13/13 & 230 & 190 & 6 & 426 & 6.205 \\
Explosion\_0 & 1 & 19/19 & 752 & 444 & 0 & 1,196 & 15.626 \\
Explosion\_1 & 1 & 64/64 & 2,592 & 1,644 & 68 & 4,304 & 57.780 \\
\midrule
\rowcolor[HTML]{ECF4FF}
\textbf{Total} & \textbf{109} & \textbf{2,099/2,099} &
\textbf{79,974} & \textbf{82,424} & \textbf{43,334} &
\textbf{205,732} & \textbf{4,115.313} \\
\end{longtable}\endgroup